\documentclass[lettersize,journal]{IEEEtran}
\usepackage{amsmath,amsfonts}
\usepackage{amssymb}
\usepackage{algorithmic}
\usepackage{array}
\usepackage{caption}
\usepackage[font=normalsize,labelfont=sf,textfont=sf]{subfig}
\usepackage[colorlinks=true, allcolors=blue]{hyperref}
\usepackage{textcomp}
\usepackage{stfloats}
\usepackage{url}
\usepackage{verbatim}
\usepackage{graphicx}
\usepackage{booktabs}
\usepackage{multirow}
\usepackage{array}
\usepackage{algorithm} 
\usepackage{mathtools}
\usepackage{bbm}
\usepackage{lipsum}
\usepackage{booktabs }
\usepackage{pifont}

\DeclareMathOperator{\diag}{diag}

\newcount\Comments  
\usepackage{color}
\definecolor{darkgreen}{rgb}{0,0.5,0}
\definecolor{purple}{rgb}{1,0,1}
\usepackage{amsmath}

\newcommand{\kibitz}[2]{\ifnum\Comments=0\textcolor{#1}{#2}\fi}

\usepackage{makecell}
\def\BibTeX{{\rm B\kern-.05em{\sc i\kern-.025em b}\kern-.08em
    T\kern-.1667em\lower.7ex\hbox{E}\kern-.125emX}}
\usepackage{balance}

\begin{document}
\title{Toward Safe Integration of UAM in Terminal Airspace: UAM Route Feasibility Assessment using Probabilistic Aircraft Trajectory Prediction}

\author{Jungwoo Cho and Seongjin Choi \thanks{J. Cho is with the Department of Air Transportation, Korea Transport Institute, South Korea. (e-mail: \href{mailto:jungwoo@koti.re.kr}{jungwoo@koti.re.kr})

S. Choi is with the Department of Civil, Environmental, and Geo- Engineering, University of Minnesota, Minneapolis, USA. (e-mail: \href{mailto:chois@umn.edu}{chois@umn.edu})

}
}

\markboth{Journal of \LaTeX\ Class Files,~Vol.~xx, No.~x, September~20xx}%
{How to Use the IEEEtran \LaTeX \ Templates}

\maketitle

\begin{abstract}
Integrating Urban Air Mobility (UAM) into airspace managed by Air Traffic Control (ATC) poses significant challenges, particularly in congested terminal environments. This study proposes a framework to assess the feasibility of UAM route integration using probabilistic aircraft trajectory prediction. By leveraging conditional Normalizing Flows, the framework predicts short-term trajectory distributions of conventional aircraft, enabling UAM vehicles to dynamically adjust speeds and maintain safe separations. The methodology was applied to airspace over Seoul metropolitan area, encompassing interactions between UAM and conventional traffic at multiple altitudes and lanes. The results reveal that different physical locations of lanes and routes experience varying interaction patterns and encounter dynamics. For instance, Lane 1 at lower altitudes (1,500 ft and 2,000 ft) exhibited minimal interactions with conventional aircraft, resulting in the largest separations and the most stable delay proportions. In contrast, Lane 4 near the airport experienced more frequent and complex interactions due to its proximity to departing traffic. The limited trajectory data for departing aircraft in this region occasionally led to tighter separations and increased operational challenges. This study underscores the potential of predictive modeling in facilitating UAM integration while highlighting critical trade-offs between safety and efficiency. The findings contribute to refining airspace management strategies and offer insights for scaling UAM operations in complex urban environments.

\end{abstract}

\section{Introduction}\label{sec:intro}
Urban Air Mobility (UAM) offers a promising solution for improving urban travel efficiency and reducing road congestion. However, integrating UAM operations into existing Air Traffic Management (ATM) systems and ensuring their interoperability presents substantial challenges, particularly in urban airspace already congested with conventional air traffic managed by Air Traffic Control (ATC). These challenges require significant modifications to current airspace management practices to ensure safe and efficient operations \cite{NASA2023roadmap}.

Moreover, as UAM technology continues to advance and the number of UAM aircraft in urban airspace increases, achieving interoperability with existing ATM systems will become increasingly complex. A potential approach to address this complexity is the allocation of exclusionary airspace specifically for UAM \cite{NASA2023roadmap}. By confining UAM activities within the designated zones, this approach is expected to reduce the need for continuous coordination between UAM and conventional aircraft. However, while aiming to simplify airspace management, it also introduces challenges such as reduced flexibility for UAM routes and the potential for fragmentation of usable airspace \cite{vascik2020geometric}. 

 
For example, an airspace assessment of the San Francisco Bay Area highlighted the limitations of exclusionary airspace under various scenarios \cite{vascik2020geometric}. The study evaluated airspace configurations from restrictive to permissive and found that in the most restrictive case, UAM operations were confined to fragmented airspace below 1,500 ft, with over two-thirds of higher-altitude airspace reserved for ATC procedures. This fragmentation resulted in longer routes, increased detours, and reduced efficiency. Even in less restrictive scenarios, where over 80\% of the airspace at or below 1,500 ft was accessible, fragmentation persisted, requiring UAM to navigate suboptimal routes and limiting operational efficiency. 

Given the limitations of exclusionary airspace, the concept of shared airspace offers a promising alternative, allowing UAM and ATM operations to coexist under strict safety and interoperability requirements. NASA's UAM airspace research roadmap \cite{NASA2023roadmap} outlines a gradual transition toward this coexistence, highlighting that Providers of Services for UAM shall coordinate access to shared airspace with ATC in higher UAM Maturity Levels (UML-4 and beyond). However, such a transition may introduce additional complexities, as it requires managing the dynamic and often unpredictable interactions between UAM and conventional aircraft, particularly in terminal airspace environments. 

NASA's X4 simulations in the Dallas-Fort Worth (DFW) region provide further insights into the challenges of integrating UAM into shared airspace \cite{Cheng2022,lee2022uamsep}. In these studies, predefined UAM corridors were designed to minimize conflicts with IFR and VFR traffic. However, Loss of Separation (LoS) incidents were observed at concerning rates, occurring once every 40 hours under a 3,000 ft horizontal and 500 ft vertical separation standard and once every 8.6 hours under the less stringent VFR standard \cite{lee2022uamsep}. These findings emphasize the complexity of maintaining safe operations in terminal airspace environments, where dense traffic and overlapping operational needs increase the risk of conflicts.

To enable coexistence between UAM and conventional air traffic, it is essential to carefully designate airspace for UAM operations and anticipate potential interactions with conventional aircraft. This involves identifying feasible UAM routes while accurately predicting conventional aircraft trajectories to maintain safe separations. However, such predictions are inherently complex due to the uncertainties introduced by dynamic air traffic patterns, environmental conditions, and human decision-making. These challenges highlight the need for robust modeling approaches capable of accounting for a wide range of potential deviations from expected flight paths.

To address these challenges, we propose a framework for assessing UAM route feasibility within terminal airspace using probabilistic trajectory prediction. The framework employs Normalizing Flows \cite{dinh2016density,scholler2021flomo}, a widely used deep generative modeling approach, to predict the short-term trajectory distribution of conventional aircraft surrounding the ego-UAM. Using these predictions, the framework evaluates UAM routes that intersect with predicted aircraft trajectories, allowing UAM vehicles to adjust their speed and avoid potential LoS events. Together, these metrics provide a structured basis for assessing UAM integration, establishing the framework as a foundation for advancing airspace management strategies that support coexistence between UAM and conventional air traffic.

The structure of the rest of this paper is as follows. Section \ref{sec:review} provides an overview of UAM airspace management and deep learning-based trajectory prediction. Section \ref{sec:method} describes the proposed framework for UAM route feasibility assessment. Section \ref{sec:result} presents the application outcomes in Seoul airspace, and Section \ref{sec:conc} summarizes the contributions and limitations of this study.

\section{Literature Review}\label{sec:review}
\subsection{UAM Airspace Management}
The successful integration of UAM into existing airspace systems requires seamless interoperability with ATM. NASA's UAM Airspace Research Roadmap highlights ATM operability as a cornerstone for future UAM systems, emphasizing the need for coordination between UAM and conventional air traffic \cite{NASA2023roadmap}. This operability would be achieved progressively through the UAM Maturity Level (UML) framework, which outlines a structured progression of operational, technical, and regulatory capabilities. At early UMLs, such as UML-1 and UML-2, operations are confined to controlled corridors with minimal interaction with ATM systems. By UML-4, UAM operations are envisioned to coordinate airspace allocation actions with ATC when necessary \cite{NASA2023roadmap}. Beyond UML-4, the roadmap anticipates highly automated systems capable of managing high-density traffic scenarios\cite{goodrich2021description}, necessitating dynamic airspace allocation strategies and collaborative traffic management. 

As UAM operations advance to higher UML levels, analyzing inherent traffic interactions becomes crucial for addressing the challenges of integrating UAM into shared airspace. A few studies have examined these challenges by simplifying the scenarios to exclude tactical UAM maneuvering. For instance, \cite{lee2022uamsep,Cheng2022} conducted a detailed analysis in high-density airspace within the Dallas-Fort Worth (DFW) region, simulating traffic demands that included scenarios with up to 1,000 UAM operations per day. The analysis revealed high LoS occurrences, particularly during enroute phases, underscoring the inherent risks in airspace environments with overlapping traffic demands. To address the risks associated with frequent and complex LoS events, exploring the role of basic UAM maneuvers, such as minor trajectory or speed adjustments, becomes crucial for maintaining safety and operational efficiency. 

The objective of this paper aligns closely with these challenges by proposing a framework that evaluates UAM route feasibility near terminal airspace using probabilistic trajectory prediction. By incorporating speed adjustments as a tactical maneuver, the proposed framework explores the feasibility of UAM operations in maintaining required safety margins. 

\subsection{Trajectory Prediction}
In our study framework, accurate trajectory prediction is essential for reducing collision risks and maintaining the necessary separation between UAM and conventional aircraft. While aircraft typically adhere to predefined flight paths, deviations can occur, especially near terminal airspace, due to environmental conditions, air traffic situations, or human factors. These uncertainties make trajectory prediction challenging, as it must account for a wide range of potential deviations and dynamic conditions.

A promising solution to this challenge is to use a data-driven deep learning approach to predict future trajectories. While some studies have explored deep learning for aircraft trajectory prediction \cite{zhao2019aircraft,zeng2020deep,ma2019aircraft}, most have focused on the \textit{deterministic} prediction, which produces a single, fixed trajectory as the output. This deterministic nature limits their ability to forecast diverse scenarios, as they lack the flexibility to represent the full range of possible outcomes. Consequently, these methods are less reliable when it comes to dynamic and unpredictable environments such as terminal airspace. 

As a result, these methods are inherently constrained in their ability to forecast diverse scenarios. By providing only a singular prediction, deterministic approaches lack the flexibility to represent the range of possible outcomes, thereby limiting their reliability in dynamic and unpredictable airspace environments. 

To address these limitations, probabilistic trajectory prediction using Deep Generative Models (DGMs) presents a compelling alternative. Unlike deterministic methods, DGMs can learn the distribution of feasible future trajectories from large amounts of data and can generate realistic samples of possible future trajectories. Ideally, using DGMs allows for the identification of the most likely trajectory, as well as the uncertainty around it.  
However, directly learning the underlying distribution from the data has been regarded as infeasible due to its high computational complexity. As a result, most existing DGMs take an indirect approach by either focusing on the ability to generate realistic samples by using adversarial training like Generative Adversarial Networks (GAN) or approximating the distribution learning objective like Variational Autoencoder (VAE).  
%
Recent advancements in flow-based generative models, known as Normalizing Flows, have shown promise in directly learning the underlying distribution \cite{dinh2016density, scholler2021flomo, choi2024gentle}. These models use a sequence of invertible transformations to map a simple base distribution into a complex distribution while allowing for the capture of the complexity of the data while still being computationally tractable. 

By learning the underlying distribution of the aircraft trajectories, Normalizing Flows can predict the probability distribution of future trajectories based on the historical observation of the aircraft. This probabilistic approach equips the model to account for uncertainties in aircraft movements, providing a foundation for integrating UAM into terminal airspace. With this information, UAM aircraft can dynamically adjust their speed to maintain safe separations, even in the presence of trajectory variability or unpredictability.

\section{Methodology}\label{sec:method}
To evaluate the feasibility of integrating UAM operations into airspace managed by ATC, we simulate UAM aircraft traversing regions specifically designated for the arrival and departure procedures of conventional aircraft. The approach seeks to minimize the likelihood of potential LoS events by adjusting the speed of UAM aircraft to maintain required safety margins, thus minimizing disruption to ATC procedures. 

\begin{figure}
    \centering
    \includegraphics[width=0.8\linewidth]{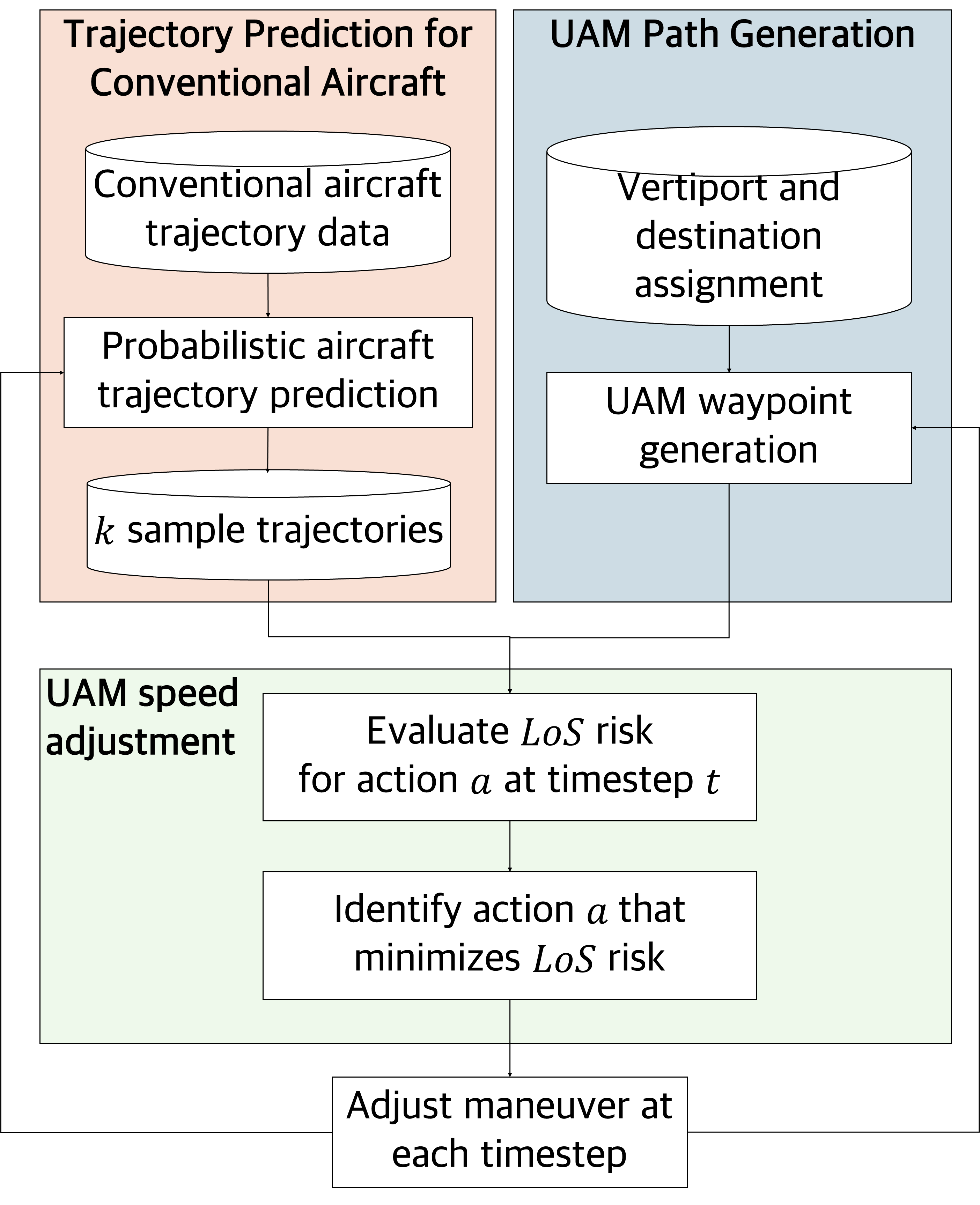}
    \caption{Block diagram of the proposed framework}
    \label{fig:framework}
\end{figure}

To achieve this, we generate predicted sample trajectories of conventional aircraft using conditional Normalizing Flows, as detailed in Section \ref{sec:nf}. These predicted trajectories provide a probabilistic representation of conventional aircraft movements and serve as the basis for evaluating potential conflicts with the planned flight path of UAM aircraft. The likelihood of LoS is assessed by comparing $k$ predicted sample trajectories of conventional aircraft with the planned UAM trajectory. If a potential LoS is detected, the UAM speed is adjusted according to the control scheme described in Figure \ref{fig:framework} and in Section \ref{sec:sa}. This scheme enables UAM operations to proceed if the predicted trajectories of conventional aircraft indicate minimal presence in the airspace.

\subsection{Scenario description}
\subsubsection{Scenarios and assumptions}
UAM flights are generated at precise 10-second intervals, departing from a vertiport and arriving at multiple designated destinations at specified altitudes. These flights follow linear paths at a constant cruising speed of 210 km/h unless a LoS event is expected. The simulation spans two days, during which the deviation of actual UAM flight times from their initially scheduled durations and analyze the separation distances between UAM and conventional aircraft at each time step. 

Trajectory prediction for conventional aircraft involves generating $k$ sets of predicted trajectories over a 60-second horizon based on preceding trajectory points sampled at 1-second intervals. These predicted trajectories are then used to assess potential convergence with UAM flight paths. A detailed description of the prediction model employed in this study is provided in Section \ref{sec:nf}.

\subsubsection{Separation standard between UAM and conventional aircraft }
Our study adopts a predefined separation threshold, which requires a horizontal separation $S_{horz}$ of 2,500 ft and a vertical separation $S_{vert}$ of 1,000 ft. UAM flight paths are designed to align with the wake turbulence advisory requirements, ensuring compliance with the specified lateral separation of 2,500 ft and vertical separation of 1,000 ft below (\cite{lee2022uamsep}).  

In addition, we define \textit{encounter} as any situation in which the separation between UAM and conventional aircraft falls below 2 nautical miles horizontally and less than 1,200 ft vertically at least one time step, as described in \cite{lee2022uamsep}. When the encounter criteria are met, further analysis determines whether the situation could lead to a potential LoS. For each encounter pair, the Closest Point of Approach (CPA) is calculated over the subsequent 60 seconds, incorporating the planned flight path of UAM and the predicted trajectory of conventional aircraft. If the CPA is projected to fall below the separation threshold, speed adjustment is applied. Details of the speed adjustment process are provided in Section  \ref{sec:sa}.

\subsubsection{Study Area Description}
\begin{figure}
    \centering
    \includegraphics[width=\linewidth]{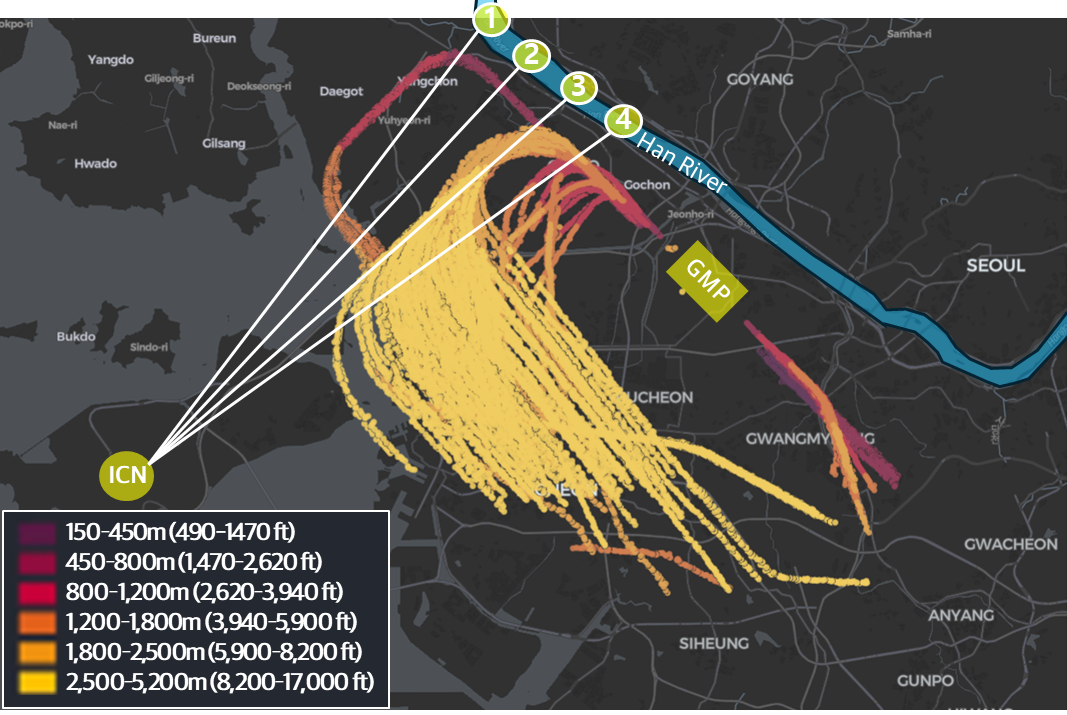}
    \caption{Illustration of the Seoul Metropolitan airspace showing UAM routes (lanes) under evaluation (white) and historic trajectories of conventional aircraft (yellow to red)}
    \label{fig:study_area}
\end{figure}

Our study area encompasses airspace over the western section of the Seoul Metropolitan area, which includes two of South Korea's busiest international airports, Incheon Airport (ICN) and Gimpo Airport (GMP). In 2019, ICN accommodated 71 million passengers and 404,104 flights. If successfully adopted, UAM could become an appealing transit option for passengers traveling between ICN and Seoul, potentially reducing travel time significantly. Meanwhile, GMP, located between ICN and central Seoul, handled more than 25 million passengers and 140,422 flights in the same year. The airspace around GMP is heavily utilized by airport arrivals and departures, as depicted in Figure \ref{fig:study_area}, presenting challenges for future UAM flights aiming to connect ICN with central Seoul. 

The GMP airport configuration consists of two runway layouts: South-East flow and North-West flow, which are adjusted based on wind conditions and noise quotas. Among the departure and approach procedures spanning diagonally from North-West to South-East, this study focuses on the North-West airspace. This sector aligns with the most direct route to the Han River, which is expected to serve as a critical corridor for UAM flights. 

As shown in Figure \ref{fig:study_area}, the evaluation focuses on routes (lanes) between ICN and four destinations along the Han River, analyzed at altitudes of 1,500 ft, 2,000 ft, 2,500 ft, and 3,000 ft to represent a range of operational scenarios. 

A segment partly connecting ICN to GMP, a 14 km route west of GMP, and an 18 km route east of GMP along the Han River have been identified as a key area for the 2025 flight demonstration, which will feature temporary vertiports placed along these routes. Studying this location is crucial for assessing the integration of UAM operations into a heavily occupied airspace.
 
\subsubsection{Efficiency and safety measures}
To assess the UAM route feasibility in terms of efficiency and safety, efficiency is measured as the deviation of actual UAM flight times from their scheduled durations. The scheduled duration is defined as the time taken for a UAM aircraft to travel from the designated vertiport, which is ICN in this case study, to one of the four designated destinations shown in Figure \ref{fig:study_area}. 

At each simulation run, a UAM aircraft traverses a specific lane at a designated altitude. During the run, speed adjustments may occur and increase flight times. Also, the minimum horizontal separation, defined as the closest horizontal distance between a UAM and a conventional aircraft when their vertical separation is less than 1,000 ft, may change due to speed adjustments. Trends in flight time deviations and minimum separations are analyzed across all runs, altitudes, and lanes, highlighting trade-offs between safety and efficiency in UAM operations in Section \ref{sec:result}.


\subsection{Probabilistic Aircraft Trajectory Prediction via Conditional Normalizing Flows}\label{sec:nf}
\subsubsection{Problem Formulation}
In this section, we briefly give out the formulation of the problem of interest. Given a time stamp $t$, we denote the location of a given aircraft as $p_t=(x_t, y_t, z_t)$, where $x_t$ represents longitude coordinate, $y_t$ represents latitude coordinate, and $z_t$ is the flight level. Given $H$ historical observations, $X_t = \left[ p_{t-H+1}, \cdots, p_t \right]$, we aim to predict the conditional probability distribution of $T$ future locations $P(Y_t|X_t)$, where $Y_t = \left[ p_{t+1} , \cdots, p_{t+T} \right]$.

\subsubsection{Density Estimation via Normalizing Flows}
Conventional generative models like Generative Adversarial Networks (GAN) and Variational Autoencoder (VAE) do not explicitly learn the likelihood of real data. On the other hand, a flow-based deep generative model learns the likelihood by using \textit{normalizing flows}. A normalizing flow describes the transformation of a probability density by using an invertible function \cite{rezende2015variational, choi2024gentle}. The idea behind using normalizing flows to learn the likelihood of the real data is that a complex distribution can be learned by sequentially applying multiple normalizing flows to a simple base distribution. For an in-depth discussion on the mathematical foundations and practical implementations of normalizing flows, we refer readers to \cite{choi2024gentle}, which provides a comprehensive overview of this topic.

For a multivariate vector $\textbf{x}$, the objective is to learn the probability density function $p (\textbf{x})$. The density $p (\textbf{x})$ is transformed into a simple distribution $p(\mathbf{z_0})$ (independent multivariate Gaussian distribution). This transformation $f$ is composed of $K$ invertible (bijective) functions $f_i, i\in \{1,...K\}$. This transformation can be denoted as follows:
\begin{equation}
    \begin{split}
    & \mathbf{x} = \mathbf{z_K} = f_K \circ f_{K-1} \circ \cdots \circ f_1(\mathbf{z_0}) = f(\mathbf{z_0}), \\
    \end{split}
\end{equation}
where we denote $\mathbf{z_i}$ as latent vector after applying $i$ invertible functions ($f_1,\cdots,f_i$).
%
By definition,

\begin{equation}
    \begin{split}
    & \mathbf{z_i} = f_i(\mathbf{z_{i-1}}), \text{ thus } \mathbf{z_{i-1}} = f_i^{-1} (\mathbf{z_i}) \\
    \end{split}.
\end{equation}

The log-likelihood of probability density function of $i$-th latent vector, $p(\mathbf{z_i})$, can be calculated as follows:

\begin{equation}
    \begin{split}
    & p_i(\mathbf{z_i}) = p_{i-1} \left(f_i^{-1} (\mathbf{z_i}) \right)  \left| \mathbf{\frac{df_i^{-1}}{dz_i}} \right| \\
    & p_i(\mathbf{z_i}) = p_{i-1} \left( \mathbf{z_{i-1}} \right)  \left|  \left( \mathbf{\frac{df_i}{d\mathbf{z_{i-1}}}} \right)^{-1} \right|  \\
    & p_i(\mathbf{z_i}) = p_{i-1} \left( \mathbf{z_{i-1}} \right)  \left|  \left( \mathbf{\frac{df_i}{d\mathbf{z_{i-1}}}} \right) \right|^{-1} \\
    & \log p_i(\mathbf{z_i}) = \log p_{i-1} \left( \mathbf{z_{i-1}} \right) - \log \left|  \left( \mathbf{\frac{df_i}{d\mathbf{z_{i-1}}}} \right) \right|.  \\
    \end{split}
\end{equation}

As a results, the probability density of $\mathbf{x}$ can be calculated as follows:
\begin{equation}
    \begin{split}
    & \log p(\mathbf{x}) = \log \pi_K (\mathbf{z_K}) \\
    & = \log p_{K-1} (\mathbf{z_{K-1}}) - \log \left|  \mathbf{\frac{df_K}{dz_{K-1}}} \right|\\
    & = \log p_{K-2} (\mathbf{z_{K-2}}) - \log \left|  \mathbf{\frac{df_{K-1}}{dz_{K-2}}} \right| - \log \left| \det \mathbf{\frac{df_K}{dz_{K-1}}} \right|\\
    & = \cdots \\
    & = \log p_{0} (z_{0}) - \sum_{i=1}^K \log \left|  \mathbf{\frac{df_i}{dz_{i-1}}} \right|.
    \end{split}
\end{equation}

Finally, the objective of training for a generative model is maximizing the likelihood of a training set, $\mathcal{D}$, as follows:
\begin{equation}
    \begin{split}
    \mathcal{L} (\mathcal{D}) = - \frac{1}{|\mathcal{D}|} \sum_{x\in\mathcal{D}} \log p(\mathbf{x})
    \end{split}.
\end{equation}

In practice, there are two conditions to consider when deciding proper functions for normalizing flows. The first condition is that, by definition, the functions should be invertible. Also, computing the Jacobian determinant should be feasible since usually computing the Jacobian of functions and computing the determinant are both computationally expensive. As a result, properly defining the function $f$ is the key to using normalizing flows. 


\subsubsection{Model Structure}
The proposed model consists of two sub-modules: 1) Condition Encoder and 2) Flow-based Decoder. The condition encoder takes the historical location as input and calculates the encoded latent vector ($h_t$). Then, the flow-based decoder uses a stack of normalizing flows to transform a simple distribution $z~N(0,1)$ to $P(Y_t|X_t)$ conditioned on the encoded latent vector computed from the condition encoder.

In the Condition Encoder, before we feed the input to the module, we preprocess the given input so that it is easy to encode the features from the data. First, we subtract each position $p_t$ from the preceding point to get relative displacement at each axis, i.e., $\Delta p_t = p_t - p_{t-1}$. Then, we concatenate the absolute coordinates ($p_t$) and the relative displacement ($\Delta p_t$) as our input to the encoder. We implemented the encoder as a recurrent neural network. Using the recurrent neural networks (RNN) is standard in encoding the sequential information of a given trajectory as shown in \cite{choi2021trajgail,choi2018network,sun2021joint}. We use three layers of Gated Recurrent Units (GRU) \cite{cho2014learning} to encode $X_t$ as $h_t$.

In the Flow-based Decoder, we calculate the probability distribution of future locations conditioned on the latent vector ($h_t$) calculated from the Condition Encoder. As mentioned in Section \ref{sec:nf}, the choice of flow function is restricted due to the two constraints that 1) the chosen function should be invertible, and 2) computing the determinant of the Jacobian matrix should be feasible. There are several feasible functions from previous studies that satisfy the constraints: linear function \cite{rezende2015variational}, $1\times1$ convolution \cite{kingma2018glow}, and affine coupling layer \cite{dinh2016density,durkan2019neural}. Among these flow functions, in this study, we use an affine coupling layer, which allows fast and flexible computation. We reformulate the affine coupling layer from \cite{dinh2016density} to a conditional affine coupling layer as follows:
\begin{equation}
    \begin{split}
    & y_{1:d} = x_{1:d}, \\
    & y_{d+1:D} = x_{d+1:D} \odot \exp{\left( s(x_{1:d} \oplus h_t ) \right)} + t(x_{1:d} \oplus h_t ), \\    
    \end{split}
    \label{eq:affine}
\end{equation}
where $s$ and $t$ are scale and translation functions, which are implemented as multi-layer perceptrons (MLP) (three layers of fully-connected neural network), $\odot$ is the element-wise multiplication operator, and $\oplus$ is the concatenation operator.

As noted in \cite{dinh2016density}, the affine coupling layer is an invertible function and guarantees fast computation of the determinant of the Jacobian matrix. First, the inverse function of Equation \ref{eq:affine} is as follows:
\begin{equation}
    \begin{split}
    & x_{1:d} = y_{1:d}, \\
    & x_{d+1:D} = \left( y_{d+1:D} - t\left( y_{1:d}\oplus h_t \right)\right) \odot \exp{\left( -s\left( y_{1:d} \right) \right)}.\\
    \end{split}
\end{equation}

Also, the Jacobian matrix of this transformation is
\begin{equation}
    \begin{split}
    \frac{\partial y}{\partial x^\top} = \begin{bmatrix}
\mathbbm{I}_d & 0 \\
\frac{\partial y}{\partial x^\top} & \diag\left( \exp{\left[ s(x_{1:d}\oplus h_t) \right]} \right)
\end{bmatrix},
    \end{split}
    \label{eq:jac}
\end{equation}
\noindent
where $\diag(*)$ is the diagonal matrix whose diagonal entries correspond to the given matrix and non-diagonal entries are zero. As a result, the Jacobian matrix shown in Equation \ref{eq:jac} is a lower-triangular matrix. Therefore, the determinant of the Jacobian matrix can be calculated as follows:
\begin{equation}
    \begin{split}
    \left| \frac{\partial y}{\partial x^\top} \right| & = \prod_{i=1}^{D} \left[ \frac{\partial y}{\partial x^\top} \right]_{i,i} =\prod_{i=1}^{d} \exp{\left[s(x_{1:d}\oplus h_t)\right]_{i,i}} \\
    &=\exp{ \left( \sum_{i=1}^{d} \left[s(x_{1:d}\oplus h_t)\right]_{i,i} \right)}, \\
    \end{split}
\end{equation}
\noindent
where $[*]_{i,i}$ is $i$-th diagonal entry of the given matrix.

Finally, the model is trained by maximizing the conditional likelihood of a training dataset as follows:

\begin{equation}\label{eq:obj}
    \begin{split}
    \mathcal{L} (\mathcal{D}) = - \frac{1}{|\mathcal{D}|} \sum_{(X_t,Y_t)\in\mathcal{D}} \log P({Y_t}|{h_t}=g({X_t}))
    \end{split},
\end{equation}

\noindent
where $g(*)$ is the condition encoder, $P(Y_t|h_t)$ is calculated from flow-based decoder, and $\mathcal{D}$ is the training dataset which contains pairs of historical observation $X_t$ and future trajectory $Y_t$.


\subsection{Performance Evaluation for Probabilistic Aircraft Trajectory Prediction}
\subsubsection{Dataset}
We conduct experiments to verify the performance of the proposed model. The trajectory data was collected from May 24th, 2022, to May 31st, 2022, near airspace near Gimpo International Airport located in South Korea. The dataset consists of 3,682 aircraft trajectories for both take-off and landing cases. We first divided the trajectory dataset into three (training, validation, and testing) with a ratio of 7:1:2 (2577:368:737). Then, we augmented each data by using a fixed moving window so that the dataset contains a fixed amount of historical observation and a fixed amount of future trajectory. We used 60 data points ($H=60$), corresponding to 60 seconds, as the observation ($X_t$) and the next 60 data points ($T=60$), corresponding to 60 seconds, as the future trajectory. Therefore, we used a moving window of 120 data points. Then, we select the $X_t-Y_t$ pairs that have more than 90\% of the data points with altitudes higher than 150 m to filter out the ground operation.

As a result, we have 23,400 pairs of $X_t$ and $Y_t$ in the training dataset, 3,539 pairs in the validation dataset, and 4,395 pairs in the testing dataset. The validation dataset is used to select the best model. We train the model only based on the training dataset, and we select the model with the best log-likelihood from the validation dataset during training. All results in the following sections are based on the testing dataset.

\subsubsection{Evaluation Metrics}
We evaluate the performance of the proposed method with the following metrics:

\begin{itemize}
    \item Minimum Average Displacement Error (minADE) - The minimum average L2 distance between the predicted trajectory and the ground truth trajectory.
    \item Minimum Final Displacement Error (minFDE) - The minimum L2 distance between the endpoint of the predicted trajectory and the ground truth trajectory.
\end{itemize}

Both minADE and minFDE are widely used metrics in trajectory prediction tasks. The minADE represents the overall accuracy of the prediction since it averages the displacement error at each timestep. Moreover, in aircraft trajectory prediction, correctly predicting the final point is critical since the final point largely affects the downstream control tasks. As a result, we use minFDE to represent the accuracy of the final point prediction.
%

\subsubsection{Baseline Model}
In the baseline model, we use a similar structure for the Condition Encoder, but we change the decoder to another network structure. We jointly train the Condition Encoder with the baseline decoder together. For the baseline models, we use Mean Squared Error (MSE) loss for training, while we use negative log-likelihood loss for the proposed model. 

\paragraph{\textbf{Recurrent Neural Network}}
Previous aircraft trajectory prediction models proposed in \cite{zeng2020deep}, \cite{shafienya20224d}, and \cite{ma2020hybrid} use recurrent neural networks (RNN) as the decoder. There are typically two choices for the RNN decoder: Long-Short Term Memory (LSTM) and Gated Recurrent Units (GRU). We used the GRU decoder for this baseline. From the historical observation $X_t$, we calculate the encoding of the observation $h_t$, and $h_t$ is used as the initial hidden state of the GRU decoder. When predicting a sequence of future locations using the GRU decoder, we use the prediction from the previous unit as the input to the next unit to recursively generate location points. This model is a \textit{deterministic} model, so we use MSE to jointly train the Condition Encoder and the GRU decoder.

\paragraph{\textbf{Multi-Layer Perceptron}}
A Multi-Layer Perceptron (MLP) is a stack of fully connected layers. From the historical observation $X_t$, we calculate the encoding of the observation $h_t$, and we calculate the final prediction, $Y_t \in \mathcal{R}^{T \times 3}$, using MLP with three layers. This model is also a \textit{deterministic} model, so we use MSE to jointly train the Condition Encoder and the MLP decoder.

\subsubsection{Results}
We trained three different models for each model depending on the input configuration. We have three choices for the input variable configuration: \emph{abs}, \emph{dev}, and \emph{abs+dev}. The ``\emph{abs}'' represents that the absolute position ($p_t$) itself is used as input, while the ``\emph{dev}'' represents that the relative displacement (${\Delta p}_{t} = p_{t}-p_{t-1}$) is used as input. The ``\emph{abs+dev}'' represents that we use both absolute position and relative displacement, and we concatenate two vectors to create the input vector. We indicate this configuration in the ``Input'' column of Table \ref{tab:result}. 

The proposed Conditional Normalizing Flow (CNF) decoder achieved significantly lower minADE and minFDE compared to deterministic baselines, highlighting its effectiveness in probabilistic trajectory prediction. Unlike deterministic models, which provide a single-trajectory prediction, the CNF decoder captures the uncertainty in trajectory prediction by modeling the distribution of possible future trajectories. This allows the model to account for inherent variations in aircraft trajectories caused by factors such as weather conditions, air traffic control directives, or pilot behavior.
Among the three input configurations, the \emph{abs+dev} combination consistently delivered the best results for all models. This suggests that combining absolute positions and relative displacements provides complementary information that enhances the model's ability to predict both the overall trajectory and endpoint accurately.

%

\begin{table}[]
    \centering
    \caption{Summary of results [km]}
    \label{tab:result}
\begin{tabular}{ll|cc}\toprule\toprule
Model &  Input & minADE & minFDE \\\midrule
\midrule
GRU+GRU & abs&      290.601 & 505.587 \\
GRU+GRU & dev&      2384.697 & 2453.00 \\
GRU+GRU & abs+dev&  265.156 & 474.137 \\
\midrule
GRU+MLP & abs&      265.487 & 496.557 \\
GRU+MLP & dev&      2948.783 & 3155.595 \\
GRU+MLP & abs+dev&  264.387 & 496.8404 \\
\midrule
\textbf{GRU+CNF} & abs&      76.80 & 25.65 \\
\textbf{GRU+CNF} & dev&      69.69 & 10.43 \\
\textbf{GRU+CNF} & abs+dev&   69.06 & 11.30 \\
\bottomrule\bottomrule
\end{tabular}
\end{table}


\subsection{UAM Speed Adjustment}\label{sec:sa}
To maintain safe separations between UAM and conventional aircraft, we propose a speed adjustment algorithm that minimizes the probability of LoS. The algorithm evaluates the acceleration or deceleration rate ($a$) of UAM aircraft to maintain a required separation, given $k=100$ predicted trajectories of conventional aircraft. 

The algorithm considers three types of adjustments: acceleration ($a>0$), deceleration ($a<0$), and maintaining current speed ($a=0$). For acceleration, a fixed acceleration rate of $+0.2g$ is applied, increasing the UAM speed $v_t$ up to the maximum cruising speed $v_{max}$, which is 210 km/h. For deceleration, the rate varies between $-0.3g$ (maximum deceleration) and 0. The algorithm computes the rate so that the UAM can stop before the relative distance between the UAM and the conventional aircraft trajectories falls below the separation threshold ($S_{horz}, S_{vert}$). 

At each time step $t$, the probability of LoS, $P^{a_t}_{LoS}(t)$, is calculated for the current acceleration or deceleration rate $a_t \in A$. The LoS probability for the current action is computed as: 
\begin{equation}
P_{LoS}^{a_t}(t)=\sum_{k=1}^{100}P^{a_t}_{LoS}(t,k),
    \label{eq:PLoS}
\end{equation}
where $P^{a_t}_{LoS}(t,k)$ is the probability of a UAM aircraft violating separation thresholds given the $k$-th sample trajectory of a conventional aircraft.

The procedure for evaluating $P^{a_t}_{LoS}(t,k)$ is as follows. The planned trajectory for UAM aircraft $i$, $X_{i}(\tau)$, is generated given its speed $v_t$ and acceleration $a_t$ at time step $t$. The $k$-th predicted trajectory of conventional aircraft $j$, $X^k_j(\tau)$ is defined over the prediction interval $\tau=(t,t+60]$. At each time step, the horizontal and vertical distances between trajectories are calculated as:
\begin{equation}
D_{horz}(\tau) = d(X_i(\tau), X^k_j(\tau)), D_{vert}(\tau) = |z_i(\tau)-z^k_j(\tau)|, 
\end{equation}
where  $d(\cdot,\cdot)$ is the geodetic distance, and $z_i$ and $z_j$ represent the altitudes of the UAM and the conventional aircraft, respectively. 

A potential LoS is identified if there exists a time $\tau^*$ such that $D_{horz}(\tau^*) \leq S_{horz}$  and $D_{vert}(\tau^*)\leq S_{vert}$ within the prediction interval. If  $\tau^*$ exists for the $k$-th trajectory, then the algorithm assigns \begin{equation}
P_{LoS}^{a_t}(t,k) = \ell^k(t),
\end{equation} 
where $\ell^k_j(t)$ is the likelihood of the $k$-th sample trajectory of aircraft $j$, obtained through trajectory prediction using conditional Normalizing Flows. Otherwise, $P_{LoS}^{a_t}(t,k)=0$. 

If $P_{LoS}^{a_t}(t)>0$, indicating a potential LoS under the current acceleration or deceleration rate, the algorithm evaluates and compares the LoS probability with other rates: $+0.2g$ (acceleration), 0 (constant speed), and $(0,-0.3g]$ (deceleration). Based on this comparison, the algorithm determines the rate that minimizes the probability of LoS. 

If $P_{LoS}^{a_t}(t)=0$, indicating no potential LoS, the algorithm evaluates the current rate as follows. If $a_t \leq 0$, it evaluates the LoS probability for $a>a_t$ and updates the rate at which the LoS probability is the minimum. If $a_t>0$, the algorithm continues with the current rate, provided the speed remains within the cruising limit $v_{max}$. 

\begin{algorithm}
\caption{Speed Adjustment of UAM Aircraft}
\begin{algorithmic}

\STATE Initialize $P_{LoS}^{a_t}(t) \gets 0$ for the current rate $a_t \gets a_{t-\Delta{t}}$
\STATE Define rate set $A = \{+0.2g, 0\} \cup (0, -0.3g]$

\STATE Input UAM trajectory $X_i(\tau)$ and $k$-th predicted conventional aircraft trajectories $X^k_j(\tau)$ for $\tau \in (t, t + l]$

\STATE Compute horizontal and vertical relative distances for all $k$ sample trajectories:
\[
D_{\text{horz}}(\tau) = d(X_i(\tau), X_j^k(\tau)) \quad \text{and} \quad D_{\text{vert}}(\tau) = |z_i(\tau) - z^k_j(\tau)|
\]

\STATE Identify LoS:
\[
\tau^* \gets \min \{\tau \in (t, t + l] : D_{\text{horz}}(\tau) \leq S_{\text{horz}} \, \land \, D_{\text{vert}}(\tau) \leq S_{\text{vert}} \}
\]

\STATE Compute the likelihood of LoS for all $k$ trajectories:
\[
P_{LoS}^{a_t}(t, k) \gets 
\begin{cases} 
\ell^k_j(t), & \text{if } \tau^* \text{ exists for the $k$-th trajectory} \\ 
0, & \text{otherwise}
\end{cases}
\]
\[
P_{LoS}^{a_t}(t) \gets \sum_{k=1}^{100} P_{LoS}^{a_t}(t, k)
\]

\IF{$P_{LoS}^{a_t}(t) > 0$}
    \STATE Evaluate $P_{LoS}^a(t)$ for all $a \in A$
    \STATE Select the acceleration or deceleration rate that minimizes $P_{LoS}^a(t)$:
    \[
    a_t \gets \operatorname{argmin}_{a \in A} P_{LoS}^a(t)
    \]
\ELSE
    \IF{$a_t \leq 0$}
        \STATE Evaluate $P_{LoS}^a(t)$ for $a > a_t$ and set $a_t \gets \operatorname{argmin}_{a > a_t} P_{LoS}^a(t)$
    \ELSIF{$a_t > 0$}
        \STATE Keep $a_t$ unchanged if $v_t \leq v_{max}$
    \ENDIF
\ENDIF

\STATE Return $a_{t}$

\end{algorithmic}
\end{algorithm}

\section{Result}\label{sec:result}

\subsection{UAM Operational Feasibility Analysis}
\begin{figure}
    \centering
    \includegraphics[width=1\linewidth]{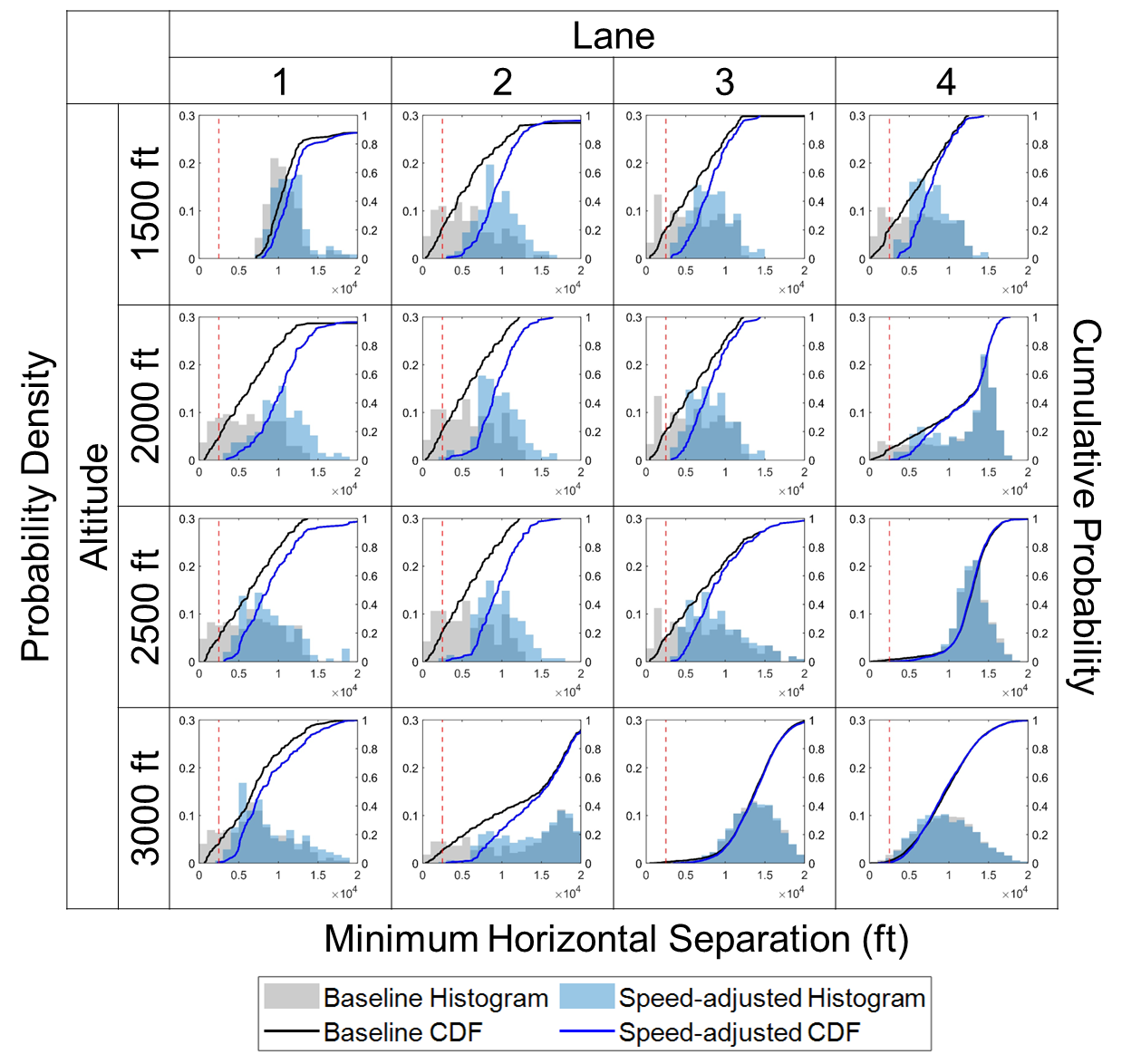}
\caption{Distribution of minimum horizontal separation for baseline and speed-adjusted scenarios}
    \label{fig:cdfs}
\end{figure}

Figure \ref{fig:cdfs} presents the histogram and cumulative distribution functions (CDFs) of minimum horizontal separations across various lane and altitude combinations under two scenarios: baseline and speed-adjusted. The baseline scenario, represented by the black lines and bars, reflects separations under conditions where no speed adjustments are applied. It serves to evaluate the inherent risks of operating in shared airspace, providing a reference for comparison with scenarios that incorporate adaptive interventions. The speed-adjusted scenario, represented by the blue lines and bars, applies dynamic speed adjustments to maintain safe separations. 

The CDFs in Figure \ref{fig:cdfs} illustrate the cumulative probabilities of achieving different minimum separations. In the baseline CDF, instances of minimum horizontal separations falling below the 2,500 ft threshold are observed in all plots except for Lane 1 at 1,500 ft. This indicates that operating UAM at this airspace inherently poses a risk of conflicts. The absence of such instances in Lane 1 at 1,500 ft suggests a relatively favorable air traffic pattern for UAM operations at this altitude and lane.

Across all lanes and altitudes, the speed-adjusted CDF consistently shifts to the right compared to the baseline CDF, reflecting improved safety margins. For example, in Lane 3 at 2,000 ft, approximately 16.9\% of cases under the baseline scenario result in separations below 2,500 ft, whereas no such cases occur in the speed-adjusted scenario. In Lane 1 at 1,500 ft, the speed-adjusted CDF also shifts slightly to the right, even though there are no cases of separations below 2,500 ft in the baseline scenario. This shift reflects the conservative nature of the proposed approach, where speed adjustments are preemptively applied based on the predicted trajectories of conventional aircraft. Specifically, if any sample trajectory is projected to result in a potential LoS, the speed adjustment mechanism modifies UAM paths conservatively to maintain a larger safety margin.

\begin{figure}
    \centering
    \includegraphics[width=1\linewidth]{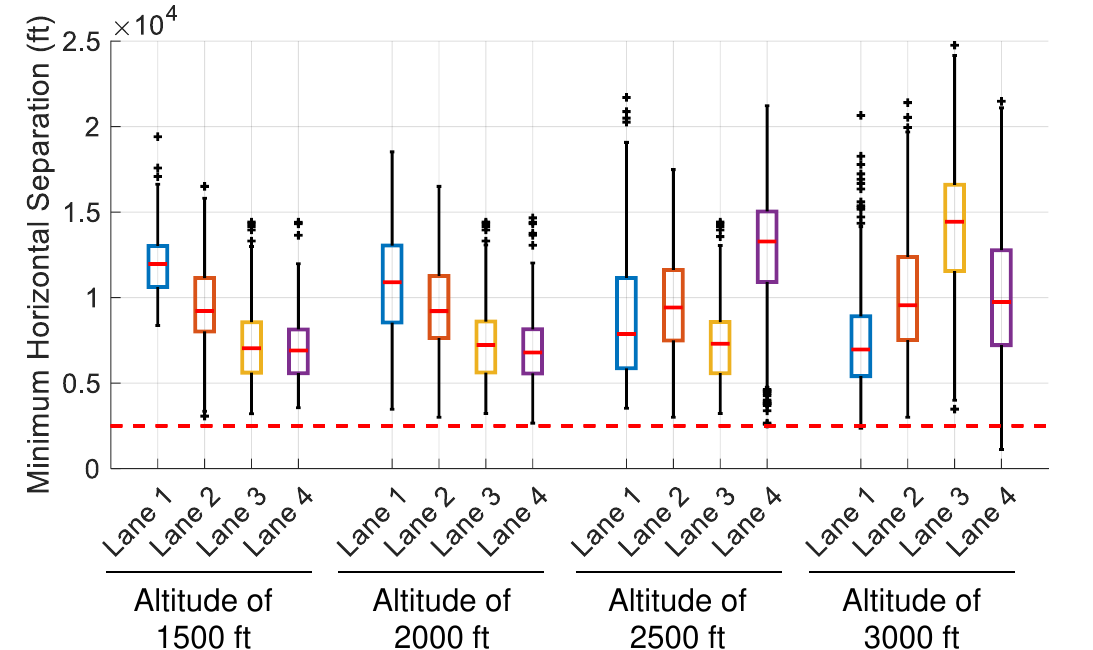}
    \caption{Minimum horizontal separation for each lane and altitude combination}
    \label{fig:boxplot_sep}
\end{figure}

The distribution of minimum horizontal separations under the speed-adjusted scenario is presented in Figure \ref{fig:boxplot_sep}, providing a more detailed analysis of variations in safety margins across different lanes and altitudes. Overall, at lower altitudes, the distributions are tighter, while at higher altitudes, they exhibit larger spreads. This highlights differences in traffic complexity, with lower altitudes offering more predictable airspace conditions and fewer interactions, while higher altitudes present more diverse interactions and dynamic traffic patterns.
 
Lane 1 at 1,500 and 2,000 ft consistently achieves the largest separations, suggesting a simpler air traffic pattern with fewer interactions between UAM and conventional aircraft. This configuration can be considered to provide UAM operations greater temporal and spatial flexibility to detect potential conflicts and adjust their trajectories proactively. Conversely, Lanes 3 and 4 at lower altitudes show smaller separations but still remain above the threshold, ensuring compliance with safety standards.

At higher altitudes, Lane 3 at 3,000 ft and Lane 4 at 2,500 ft show relatively large separations but with greater variability. The more dispersed distributions reflect increased traffic density and complex interaction scenarios between UAM and conventional aircraft, highlighting the operational challenges of these airspace configurations. 

Meanwhile, nearly all data points exceed the 2,500 ft threshold, with only a few outliers observed at 3,000 ft altitude. Outliers below the 2,500 ft threshold, primarily observed at 3,000 ft, represent instances where separations approach the critical safety limit. These cases, though rare, may arise from highly constrained scenarios requiring further investigation.

\begin{figure}
    \centering
    \includegraphics[width=1\linewidth]{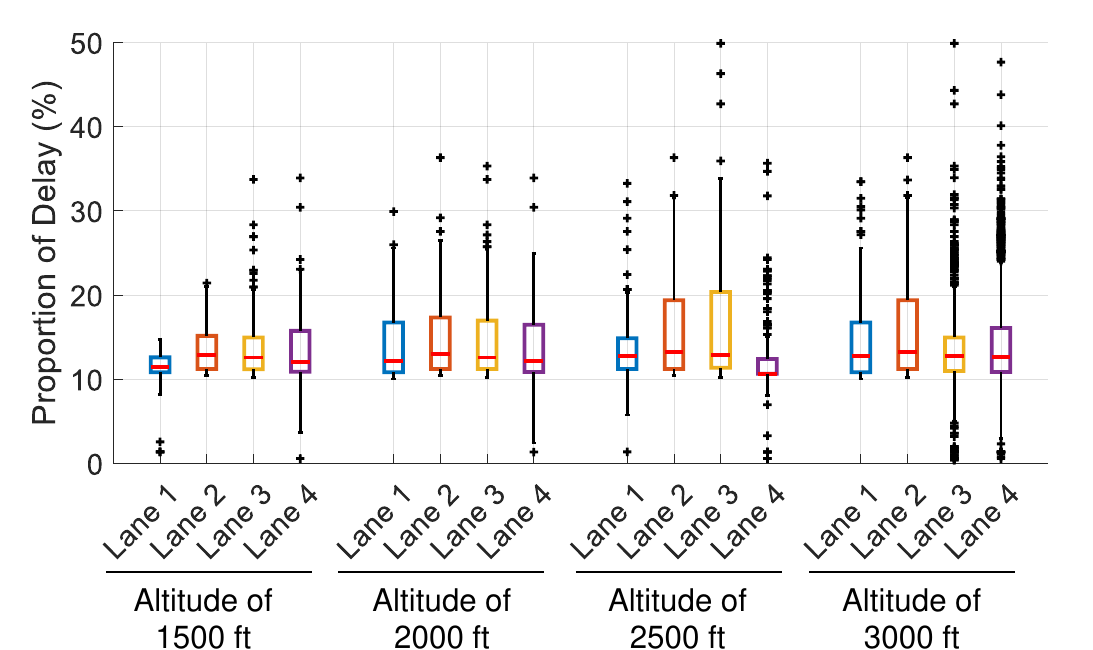}
    \caption{Proportion of flight delay for each lane and altitude combination}
    \label{fig:boxplot_delay}
\end{figure}

The boxplots in Figure \ref{fig:boxplot_delay} depict the proportion of flight delay, calculated as the delayed time divided by the initial planned time, across various lane and altitude combinations. Median delay proportions are relatively consistent, ranging between 10.4\% and 13.6\%, but the variability, represented by the spread of delay proportions, differs across lanes and altitudes.

Lane 1 at 1,500 ft exhibits the least variability, with narrow IQRs and fewer outliers. This suggests that speed adjustments, in this case, are less disruptive, resulting in predictable and stable delays. In contrast, Lanes 2 and 3 show greater variability, particularly at altitudes of 2,500 ft and 3,000 ft. While the median delay in these lanes remains close to 12\%, individual cases occasionally experience significantly higher delays due to encounters with multiple aircraft in sequence.

Lane 4 at 2,500 ft shows a relatively small spread but includes occasional high-delay occurrences, likely due to multiple encounters with conventional aircraft requiring more frequent decelerations. At 3,000 ft, Lane 4 shows substantial variability, reflecting the influence of diverse and challenging traffic conditions. The higher variability highlights the greater traffic density and operational complexity at higher altitudes, necessitating more extensive adjustments.

In conclusion, Lane 1 at lower altitudes emerges as a favorable option when considering both safety and efficiency. It consistently achieves the largest minimum horizontal separations with minimal variability, ensuring compliance with the 2,500 ft safety threshold, and demonstrates the lowest and most stable delay proportions. Lane 4 at 2,500 ft also offers a viable alternative, with a relatively small spread, a low median delay, and large separations. Despite its frequent encounters, Lane 4's proximity to central Seoul enhances its accessibility to potential vertiport locations concentrated in urban areas.

\begin{figure}
    \centering
    \includegraphics[width=1\linewidth]{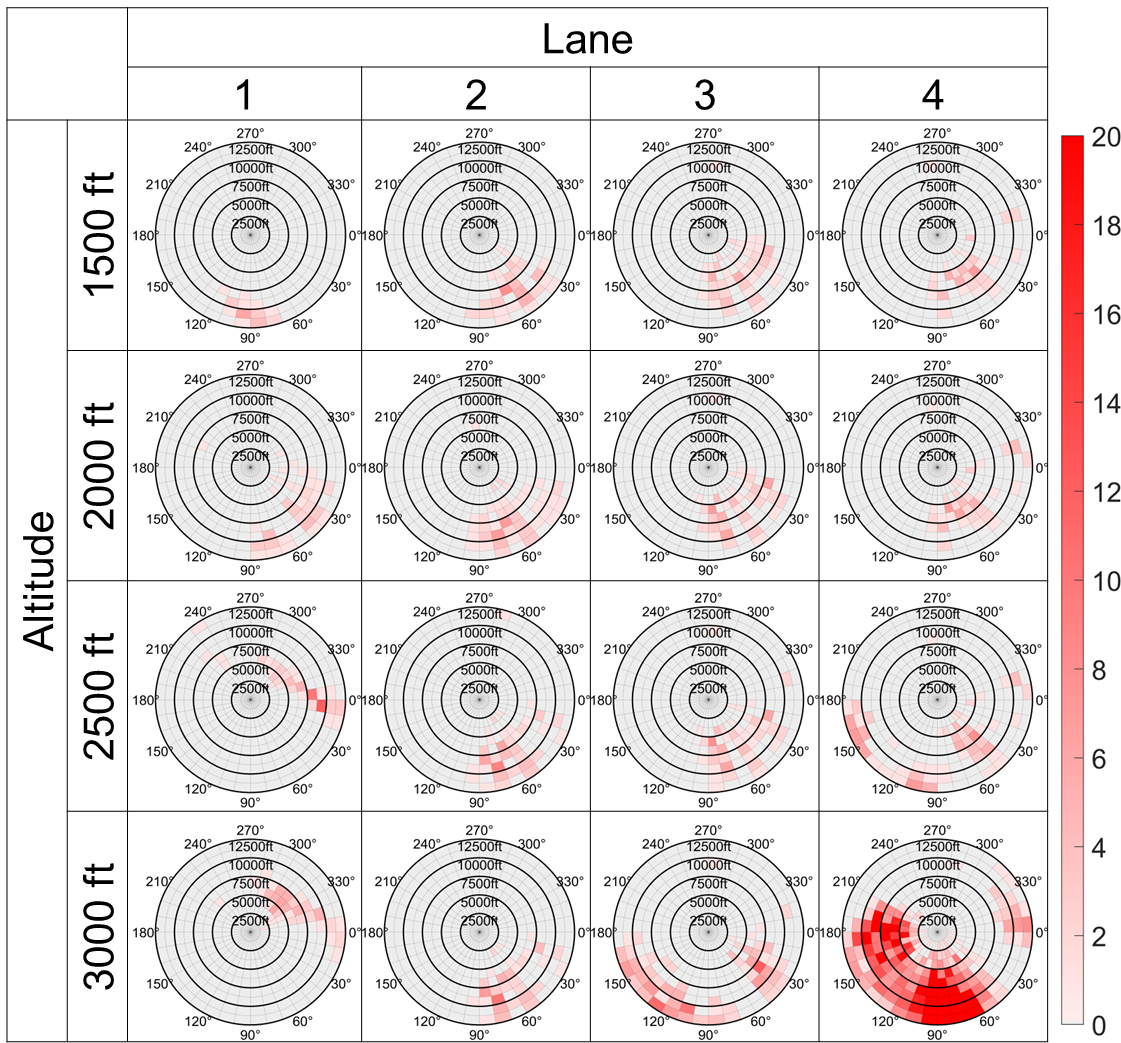}
    \caption{Polar heatmap of relative bearing and relative horizontal separation at the closest point of approach between UAM and conventional aircraft}
    \label{fig:rel_bearing}
\end{figure}
Figure \ref{fig:rel_bearing} presents the frequency of relative positions between UAM and conventional aircraft in polar coordinates at their CPA. The angular axis represents the relative bearing, while the radial axis shows the horizontal separation at the CPA, with the innermost circle marking the 2,500 ft threshold. Since this plot reflects the CPA, the separation increases in subsequent time steps, indicating that the UAM and conventional aircraft diverge after the closest approach.

A cluster of points between 0° and 90° indicates CPAs with conventional aircraft on the right side of the UAM, primarily associated with departing traffic from the airport. Similarly, CPAs between 270° and 360° reflect interactions with intruders on the left side of the UAM, often representing arriving aircraft at higher altitudes, particularly in Lane 1. CPAs with bearings between 90° and 270° indicate that the conventional aircraft is necessarily in the rear sector of the UAM and is moving away, confirming successful divergence post-CPA.
 
The distribution of relative bearing and separation differ across lanes. Lane 1 exhibits sparse distributions that are generally concentrated in the outer radial regions at lower altitudes, indicating that encounters occur predominantly at larger horizontal distances. At higher altitudes, this lane is more associated with arrival traffic, resulting in CPAs in closer radial bands. Despite this, the configuration supports smoother UAM operations, as the lower air traffic density and relative isolation from busy flight paths reduce the frequency and complexity of required adjustments to maintain safe separations.  

Lanes 2 and 3 show CPA distributions mostly in the intermediate radial bands, particularly in certain angular regions such as 0°–90° and 270°–360°, with only a few instances falling below the 5,000 ft threshold. The interactions are primarily linked to departing traffic from the airport, and the available room for adaptive speed changes in these lanes remains sufficient to avoid LoS.

\begin{figure}
     \centering
     \includegraphics[width=1\linewidth]{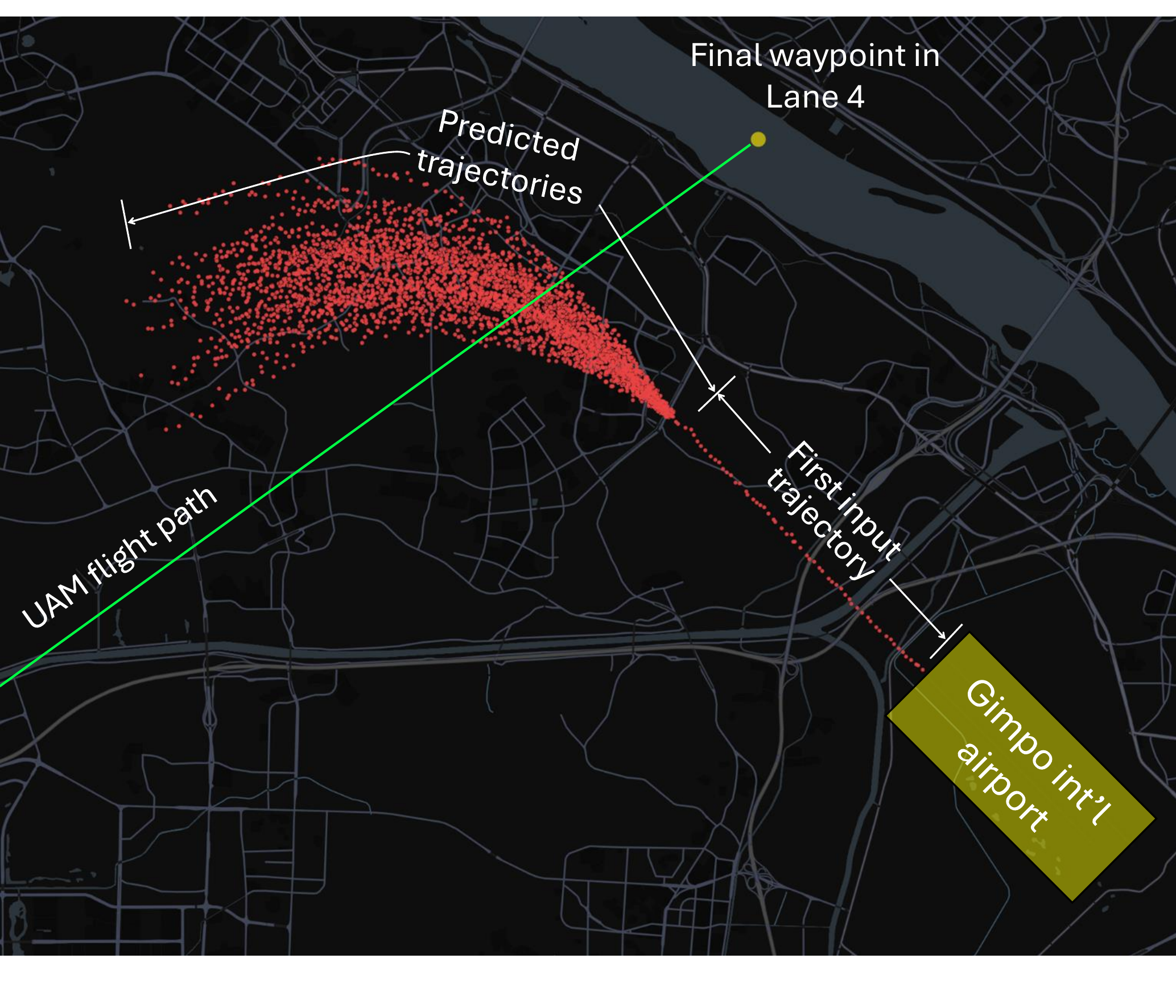}
     \caption{Case illustration of UAM flight path (green) intersecting with predicted aircraft trajectories (red) generated from the first input data}
     \label{fig:los_case}
 \end{figure}

Lane 4 exhibits occasional points near or within the 2,500 ft threshold. This lane, being closest to the airport, frequently overlaps with departing aircraft. Due to its proximity to the airport, these departing aircraft often have insufficient historical trajectory data within the 60-second prediction window to produce forward predictions. As a result, UAMs operating in this lane may lack actionable prediction results in time to preemptively adjust their speed. This limitation, as illustrated in Figure \ref{fig:los_case}, leads to cases where UAMs are unexpectedly presented with predicted trajectories of aircraft expected to cause a LoS within only a few seconds. 

\section{Conclusion}\label{sec:conc}

Our framework provides a preliminary feasibility assessment for integrating UAM routes within airspace managed by ATC operations. Instead of focusing on specific combinations of altitudes and lanes for operational deployment, this study explores the viability of this approach, while examining the inherent trade-offs between safety and efficiency.

While this study demonstrates the feasibility of integrating UAM operations using a predictive framework for route evaluation, certain limitations should be noted. The analysis was based on a week-long dataset, which, although sufficient for demonstrating initial feasibility, limits the ability to fully capture the variability of air traffic patterns. Incorporating a larger and more diverse dataset could enhance the robustness and accuracy of predictive models. 

Furthermore, this study employed a simplified approach to UAM maneuvering, focusing exclusively on adjusting acceleration rates while excluding more complex maneuvers such as altitude changes or heading adjustments. However, it should be noted that this simplification does not capture the full maneuvering capabilities of eVTOL aircraft, which are designed to perform advanced adjustments in altitude, heading, and speed to navigate complex airspace scenarios. By narrowing the scope of maneuvering, the study aims to provide a baseline assessment of UAM integration feasibility, leaving the evaluation of more sophisticated control strategies for future research.

Lastly, the framework was tested primarily in a single-airport scenario. While conceptually applicable to regions with multiple airports and more complex traffic patterns, further research is needed to validate its performance and scalability in these contexts. Addressing these limitations in future studies would refine the framework and broaden its applicability in diverse operational environments. 


\bibliographystyle{ieeetr}
\bibliography{sample}

\newpage

\vskip 0pt plus -1fil
\begin{IEEEbiography}[{\includegraphics[width=1in,height=1.25in,clip,keepaspectratio]{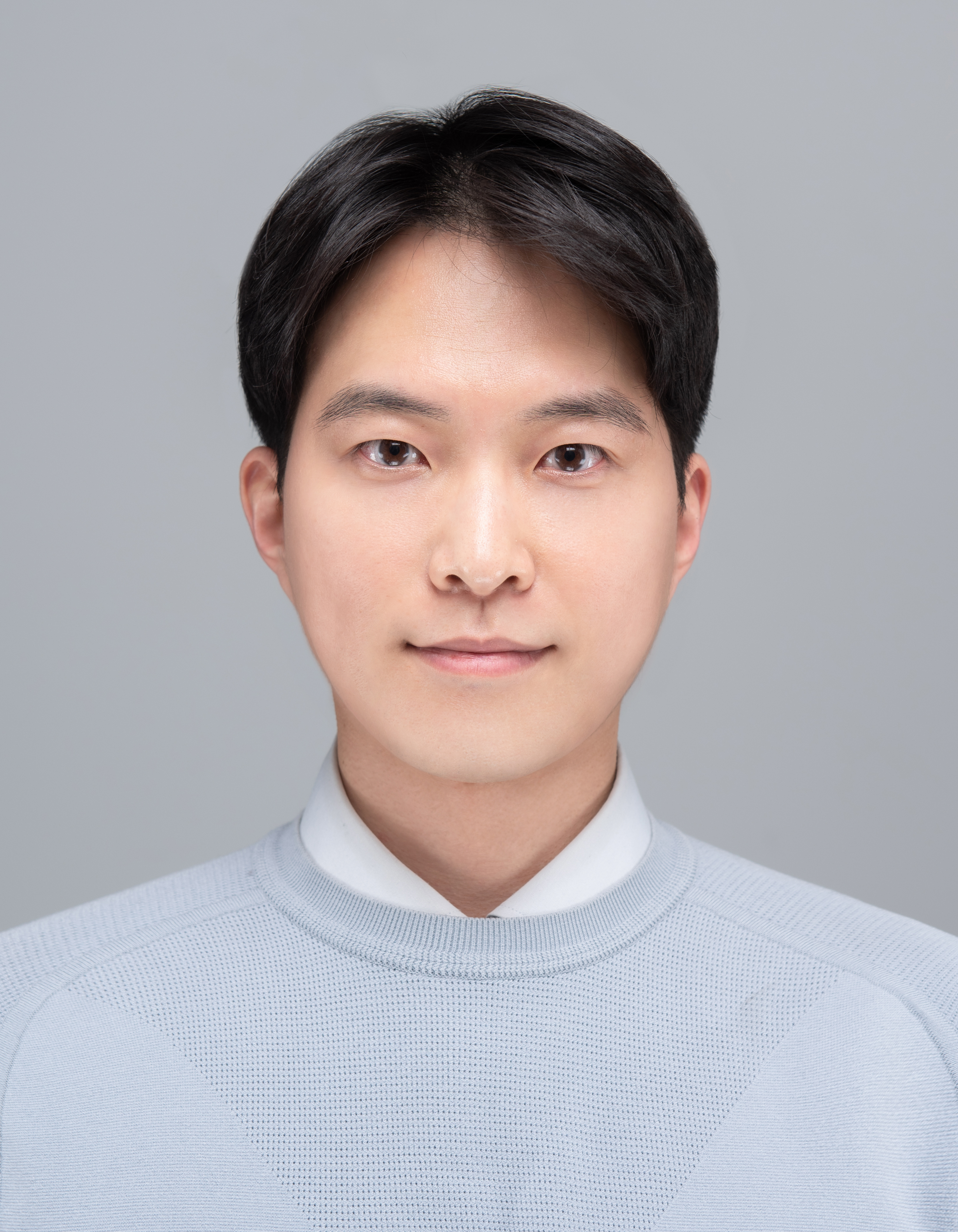}}]{Jungwoo Cho} received B.S., M.S., and Ph.D. degree in the Department of Civil and Environmental Engineering from Korea Advanced Institute of Science and Technology, Daejeon, Republic of Korea, in 2015, 2017 and 2021, respectively. He is currently an associate research fellow at Korea Transport Institute, Sejong, Republic of Korea. His current research centers on UAM Airspace Design, Air Traffic Flow Management, and Aviation Data Analytics.\end{IEEEbiography}
\vskip 0pt plus -1fil
\begin{IEEEbiography}[{\includegraphics[width=1in,height=1.25in,clip,keepaspectratio]{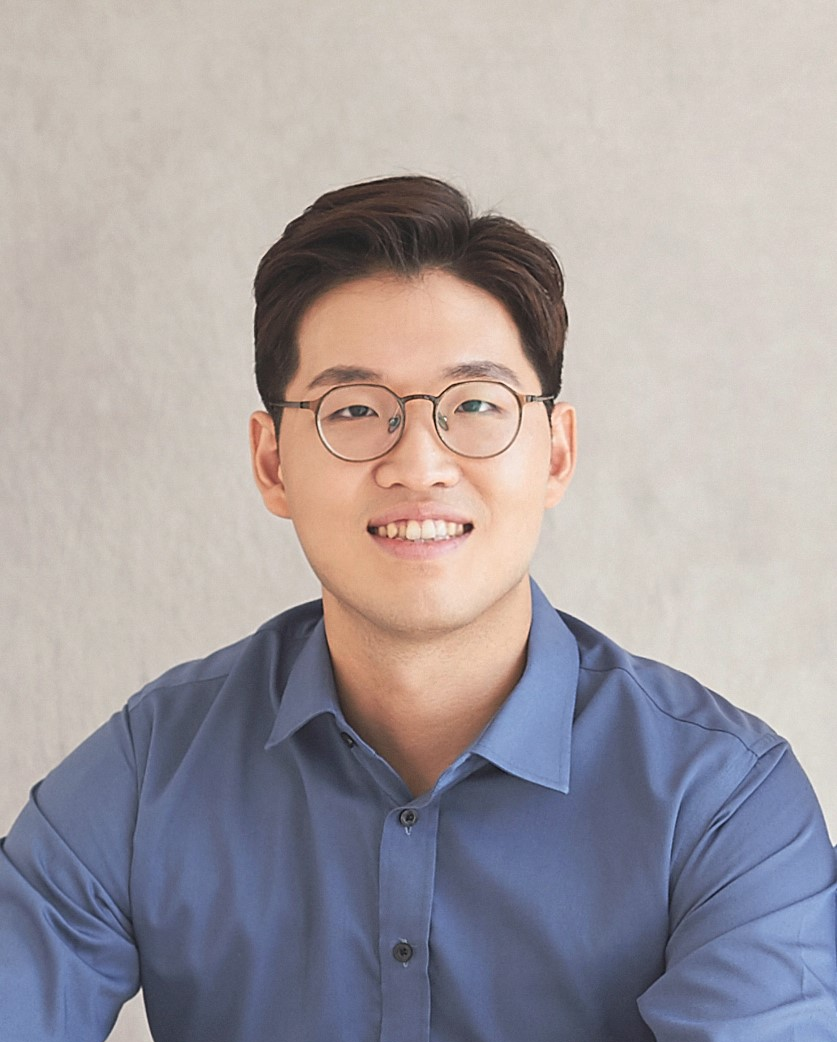}}]{Seongjin Choi} received the B.S., M.S., and Ph.D. degree in the Department of Civil and Environmental Engineering from Korea Advanced Institute of Science and Technology, Daejeon, Republic of Korea, in 2015, 2017 and 2021, respectively. He is currently an Assistant Professor at the Department of Civil, Environmental, and Geo- Engineering at the University of Minnesota, Minneapolis, USA. His research centers on Urban Mobility Data Analytics, Spatiotemporal Data Modeling, Deep Learning and Artificial Intelligence, and Connected Automated Vehicles and Cooperative ITS.
\end{IEEEbiography}

\end{document}